\documentclass{article}

    \PassOptionsToPackage{numbers}{natbib}
    \usepackage[preprint]{neurips_2019}

\usepackage[utf8]{inputenc} % allow utf-8 input
\usepackage[T1]{fontenc}    % use 8-bit T1 fonts
\usepackage{hyperref}       % hyperlinks
\usepackage{url}            % simple URL typesetting
\usepackage{booktabs}       % professional-quality tables
\usepackage{amsfonts}       % blackboard math symbols
\usepackage{nicefrac}       % compact symbols for 1/2, etc.
\usepackage{microtype}      % microtypography
\usepackage[pdftex]{graphicx}
\usepackage{subfig}
\usepackage{multirow}
\usepackage{wrapfig,lipsum,booktabs}
\usepackage{mathtools}
\usepackage{float}

\title{BasisConv: A method for compressed representation and learning in CNNs}

\author{%
  Muhammad Tayyab\thanks{http://www.mtayyab.com} \\
  Center for Research in Computer Vision\\
  Department of Computer Science\\
  University of Central Florida, USA\\
  \texttt{tayyab@knights.ucf.edu} \\
   \And
   Abhijit Mahalanobis \\
  Center for Research in Computer Vision\\
  Department of Computer Science\\
  University of Central Florida, USA\\
  \texttt{amahalan@crcv.ucf.edu} \\
}

\begin{document}

\maketitle

\begin{abstract}
It is well known that Convolutional Neural Networks (CNNs) have significant redundancy in their filter weights. Various methods have been proposed in the literature to compress trained CNNs. These include techniques like pruning weights, filter quantization and representing filters in terms of a basis functions. Our approach falls in this latter class of strategies, but is distinct in that that we show both compressed learning and representation can be achieved without significant modifications of popular CNN architectures. Specifically, any convolution layer of the CNN is easily replaced by two successive convolution layers: the first is a set of \textit{fixed} filters (that represent the knowledge space of the entire layer and do not change), which is followed by a layer of one-dimensional filters (that represent the learned knowledge in this space). For the pre-trained networks, the fixed layer is just the truncated eigen-decompositions of the original filters. The 1D filters are initialized as the weights of linear combination, \textit{but are fine-tuned to recover any performance loss due to the truncation}. For training networks from scratch, we use a set of \textit{random orthogonal fixed filters} (that never change), and learn the 1D weight vector directly from the labeled data. Our method substantially reduces i) the number of learnable parameters during training, and ii) the number of multiplication operations and filter storage requirements during implementation. It does so without requiring any special operators in the  convolution layer, and extends to all known popular CNN architectures. We demonstrate the generality of the proposed approach by applying it to four well known network architectures with three different data sets. The results show a consistent reduction in i) the number of operations by up to a factor of 5, and ii) number of learnable parameters by up to a factor of 18, with less than 3\% drop in performance on the CIFAR100 dataset.
\end{abstract}

\section{Introduction}

While there has been a tremendous surge in convolutional neural networks and their applications in computer vision, relatively little is understood about how information is learned and stored in the network. This is evidenced by the fact that researchers have successfully proposed different approaches for compressing a network after it has been trained \cite{Cheng2017ASO}, including techniques like pruning weights \cite{optimalbrain, Surgeon, Han2016DeepCC, Han2015LearningBW, Srinivas2015DatafreePP, Chen2015CompressingNN}, assuming row-column separability \cite{Jaderberg2014SpeedingUC}, applying low rank approximations for computational gains \cite{Denton2014ExploitingLS}, and using basis representation \cite{Jaderberg2014SpeedingUC, Qiu2018DCFNetDN} .  It is clear that CNNs do not need to explicit learning of a large number of coefficients in the manner in which they are currently trained.    Based on this observation, we take a different view of the key component in CNNs - the filtering operation - and propose a fundamentally different approach that combines a "fixed" convolution operator (that is never trained or learned) with a learnable one-dimensional kernel. This is motivated by a salient observation that the filters are points in a hyper-dimensional space that is learned via the training process.  We claim that the filters themselves are not important in the end, but it is the representation of the space itself is the key. \par
For networks that have been already trained, the underlying knowledge space of a layer can be easily represented as truncated eigen decomposition of the filters. We can then efficiently fine-tune the coefficients of linear combination to find new points in this lower dimensional space which recover any loss in performance, and discard the original filters. As we will show, this approach dramatically reduces the number of filtering operations and filter storage requirements, without notable drop in performance. The same construct can be also use to train a network from scratch without having to explicitly learn the filter kernels across the network. For this scenario, we show that random basis functions can be used as fixed convolution kernels (which never require training), with one dimensional weight vectors that learn the relevant information.  We refer to this ability to learn in a compressed format as "compressed learning" where instead of learning the 3D filter parameters, we only need to learn relatively fewer parameters that describe where these filters reside in the hyperdimensional information space in a given layer of the CNN. Thus, this paper unifies the goals of compressing previously trained networks, and  training networks in a compressed format when learning new information from scratch. \par

\begin{figure}
  \centering
  \includegraphics[clip, trim=0cm 0cm 1cm 0cm, scale=0.4, width=\textwidth]{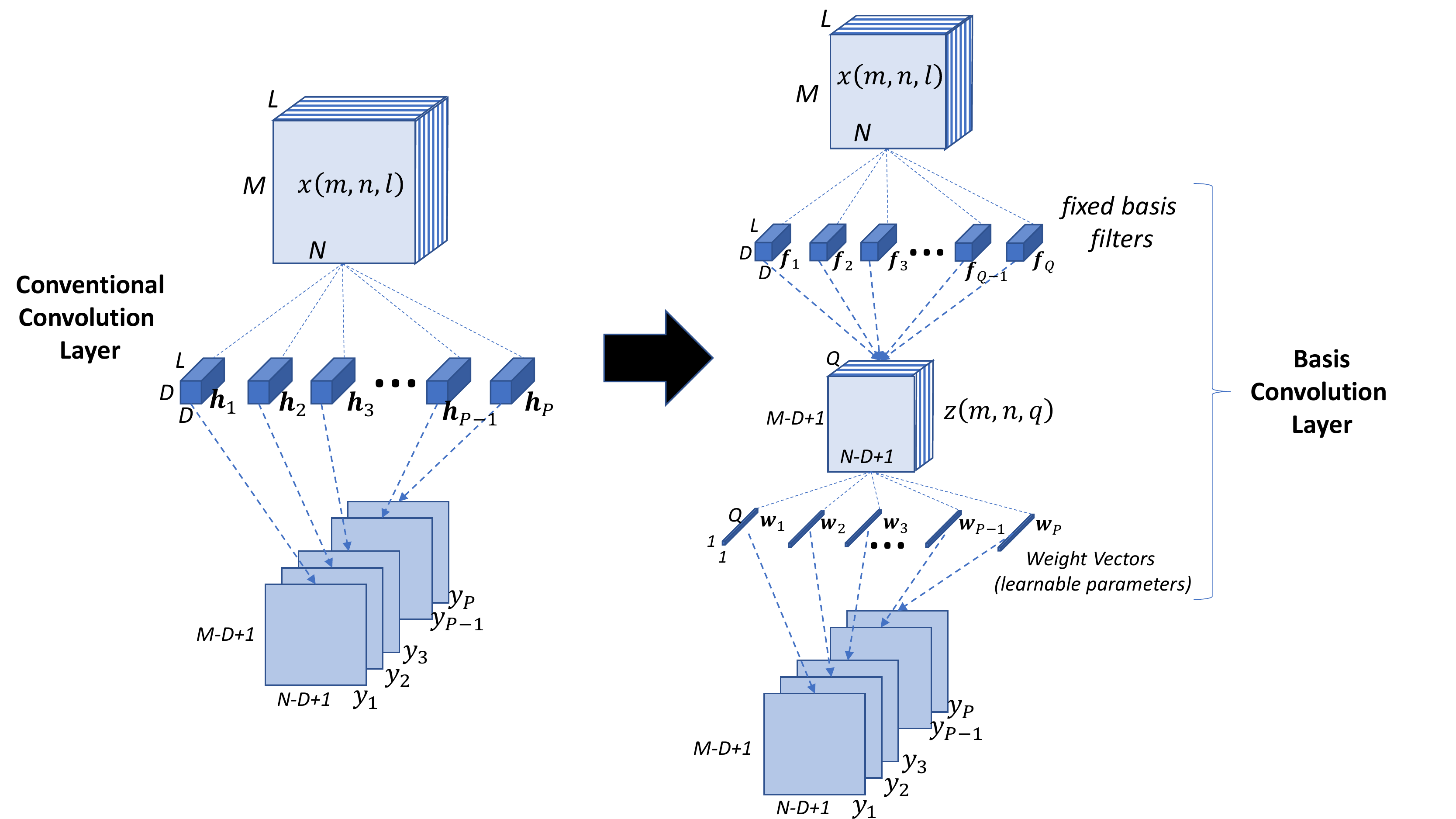}
  \caption{A side by side comparison of conventional Convolutional layer (left) and Basis Layer (right).}
  \label{fig:basisconv}
\end{figure}

\section{Compressed Representation and Learning }
Consider the fundamental convolution operation in any given layer of a convolutional neural network depicted on the left in Figure 1. Assume that an input block of data $x(m,n,l)$ (such as the activations or output of the previous layer) is convolved with a set of 3D filters $ h_k (m,n,l), \ k=1…P$. The output $y_k (m,n)$ can be expressed as

\begin{equation}
\label{eq:1}
% \small
y_k(m, n) = x(m, n, l) \ * \ h_k(m, n, l), \qquad 1 \leq k \leq p
\end{equation}

where $*$ represents the convolution operation. The right side of Figure 1 shows how the same output can be obtained using two successive convolution stages. Here, we assume that the filters can be expressed as a linear combination of Q basis functions $f_i (m,n,l), \quad i=1…Q$, such that

\begin{equation}
\label{eq:2}
% \small
h_k (m, n, l) = \sum_{i=1}^{Q} w_{ik} \cdot f_i(m, n, l)    
\end{equation}

where $w_{ik}$ are the weights of linear combination. Using this representation, the output can be expressed as 
\begin{equation}
\label{eq:3}
% \small
y_k(m, n) = \sum_{i=1}^{Q} w_{ik} \cdot [x(m, n, l) * f_i(m, n, l)], \quad 1 \leq k \leq P    
\end{equation}

The key observation is that the Q convolution terms  $z_i (m,n)=x(m,n,l)*f_i (m,n,l)$ need to be computed only once, and they are common to all $P$ outputs  $y_k (m,n)$.  These can be stacked together to form the 3D intermediate result $z(m,n,q)$ while the weights $w_{ik}$ can be treated as $1 \times 1\times Q$ filter $w_k (q)$. Therefore, the outputs $y_k (m,n)$ are simply the convolution of two, i.e 
\begin{equation}
% \small
y_k (m,n)=w_k (q)*z(m,n,q)
\end{equation}
We refer to this construct using two successive convolutions as BasisConv.
\subsection{Compression of pretrained Networks}
\label{Compression-of-pretrained-Networks}

% The resulting architecture of the \textit{basis} convolution layer is shown in Figure \ref{fig:basisconv} on the right. The key observation is that the Q convolution terms  $x(m,n,l)*f_i (m,n,l)$ need to be computed only once, and they are common to all P outputs  $y_k (m,n)$. 

% Depending on the choices of $P$ and $Q$, this can lead to substantial reduction in the number of multiplication operations. Specifically, let $Z$ represents the number of multiplications for one convolution operation (between $x(m,n,l)$ and either $h_k (m,n,l)$ or $f_i (m,n,l)$). If the size of the filters is $D \times D \times L$, and the size of the input data is $M \times N \times L$, It is easy to show that $Z=LD^2 (M-D+1)(N-D+1)$. Therefore, multiplications required in Eq. (\ref{eq:1}) is 

% \begin{equation}
% % \small
% A = PZ = PLD^2 (M-D+1)(N-D+1)
% \end{equation}
% while multiplications required in Eq (3) is

% \begin{equation}
% \begin{split}
%     B & = QLD^2 (M-D+1)(N-D+1)+PQ(M-D+1)(N-D+1) \\
%     & = Q[L^2+P](M-D+1)(N-D+1)
% \end{split}
% % \small
% \end{equation}

% We see that the ratio of the two is

% \begin{equation}
% % \small
% \frac{B}{A}= \frac{(PLD^2 (M-D+1)(N-D+1))}{Q[LD^2+P](M-D+1)(N-D+1)} = \frac{PLD^2}{Q[LD^2+P]}
% \end{equation}

% Thus, as long as $LD^2 >> P$, the number of multiplications will be reduced by a factor close to $P/Q$. 
% Since the filters contain redundant information, we can find different linear combination strategies where $Q<P$, and thereby realize a substantial decrease in the number of multiplication operation.
It is well known that eigen decomposition results in a compact basis that minimizes the reconstruction error achieved by a linear combination of basis functions. We therefore choose $f_i (m,n,l)$ as the eigen filters that represent the sub-space in which the original filters $h_k (m,n,l)$ lie. To obtain the eigen filters, we define the $LD^2 \times 1$ dimensional vector  $\mathbf{h_k}$ as a vectorized representation of  $h_k (m,n,l)$, and construct the matrix $\mathbf{A}= [\mathbf{h_1} \ \mathbf{h_2} \ ... \ \mathbf{h_P}]$ with $\mathbf{h_k}$ as its columns. The eigenvectors of $\mathbf{A A^T}$ represent the sub-space of the filters, and satisfy the relation $\mathbf{A A^T f_i}=\lambda_i \mathbf{f_i}$, where $\mathbf{f_i}$ are the eigenvectors, and $\lambda_i$ are the corresponding eigenvalues. The eigen filter  $f_i (m,n,l)$ is readily obtained by re-ordering the elements of the eigenvector $\mathbf{f_i}$ into a $D \times D \times L$ array.  Although the number of possible eigenvectors is equal to the dimensionality of the space, we select a small subset of eigenvectors which correspond to the largest Q eigen-values that best represent the dominant coordinates of the filters’ subspace. Since the eigen-values  represent the information present in each eigen-vector, in practice we will use the metric $t = \frac{ \sum_{i=1}^{Q} \lambda_i}   { \sum_{i=1}^{LD^2} \lambda_i}$ to choose Q such that most of the relevant information is retained in the selected eigen-vectors. \par
The decomposition of the filters $h_k$ of any given layer of the network can be succinctly expressed in matrix vector notation by defining $\mathbf{F}= [\mathbf{f_1} \ \mathbf{f_2} \ ... \ \mathbf{f_Q}]$  (i.e. the matrix of eigenvectors of the filters for that layer)  so that 

\begin{equation}
    \mathbf{h_k} = \mathbf{Fw_k}    
\end{equation}

and $\mathbf{w_k} = [w_{1k}\ w_{2k}\ ... \ w_{Qk}]$ is a $Q \times 1$ vector of weights. Since $\mathbf{F^T F=I}$ (i.e. the identity matrix), the weights of easily obtained by computing

\begin{equation}
    \mathbf{w_k} = \mathbf{F^T h_k}    
\end{equation}

 Depending on the choices of $P$ and $Q$, this can lead to substantial reduction in the number of multiplication operations. Specifically, let $O$ represents the number of multiplications for one convolution operation (between $x(m,n,l)$ and either $h_k (m,n,l)$ or $f_i (m,n,l)$). If the size of the filters is $D \times D \times L$, and the size of the input data is $M \times N \times L$, It is easy to show that $O=LD^2 (M-D+1)(N-D+1)$. Therefore, multiplications required in Eq. (\ref{eq:1}) is 

\begin{equation}
% \small
A = PO = PLD^2 (M-D+1)(N-D+1)
\end{equation}
while multiplications required in Eq (3) is

\begin{equation}
\begin{split}
    B & = QLD^2 (M-D+1)(N-D+1)+PQ(M-D+1)(N-D+1) \\
    & = Q[L^2+P](M-D+1)(N-D+1)
\end{split}
% \small
\end{equation}

We see that the ratio of the two is

\begin{equation}
% \small
\frac{A}{B}= \frac{PLD^2 (M-D+1)(N-D+1)}{Q[LD^2+P](M-D+1)(N-D+1)} = \frac{PLD^2}{Q[LD^2+P]}
\end{equation}

Thus, as long as $LD^2 >> P$, the number of multiplications will be reduced by a factor close to $P/Q$ (i.e. the ratio of the the original number of filters and the number of basis filters used). 

% \begin{figure}
%   \centering
%   \fbox{\includegraphics[clip, trim=2cm 0cm 3cm 0cm, scale=0.4]{Figures/eiegn-space.pdf}}
%   \caption{The original filter embeddings are used for learning the knowledge space of each layer. Finetuning moves these embedding to new locations in the same subspace which mitigates any overall loss in performance}
%   \label{fig:eiegn-space}
% \end{figure}

\subsection{compressed Learning}

The architecture shown in Figure \ref{fig:basisconv} is not only amenable to reducing the filter storage requirements and multiplications required for each convolution layer, but is amenable to learning in the compressed space where the number of \textit{learnable parameters} is substantially reduced. 
% The reason is the weight vectors $\mathbf{w_k}$ are the lower-dimensional embeddings of the original filters in the sub-space represented by the informative eigenvectors. Thus, while the eigenvectors represent “knowledge space” captured in a given layer of the network, and the weights represent specific points within this space where each filter resides. This observation allows us to fine-tune the weights directly to mitigate the approximation errors at each layer, without having to re-learn the original filter parameters.
Recall that number of learnable parameters in the original filters is $LD^2$. Since there are $P$ such filters, the total number of original learnable parameters is $PLD^2$.   However, the total number of "learnable" parameters for BasisConv is $PQ$ (depicted in Figure \ref{fig:basisconv} as a $P$ one-dimensional filters of length $Q$). Therefore, the reduction in the number of learnable parameters is $LD^2/Q$. If $LD^2>>Q$, it is clear that the number of scalar weights that need to be refined is substantially less than the original number of learnable parameters. \par

For pretrained networks, fine tuning is achieved by retraining $\mathbf{w_k}$ while freezing the eigenfilters in each basis convolution layer. The reason is the weight vectors $\mathbf{w_k}$ are the lower-dimensional embeddings of the original filters in the sub-space represented by the informative eigenvectors. Thus, while the eigenvectors represent “knowledge space” captured in a given layer of the network, the weights represent specific points within this space where each filter resides. This observation allows us to fine-tune the weights directly to mitigate the approximation errors at each layer (without having to fine-tune the filters explicitly). \par
The more interesting scenario arises for \textit{training a network from scratch} in compressed format. Of course, if the final values of  $\mathbf{h_k}$ are not known, then it is not possible to use eigen space representation. Therefore, for \textbf{\textit{compressed learning}} from scratch, we propose to initialize the columns of $\mathbf{F}$ with random vectors that will remain fixed, and only allow the coefficients $\mathbf{w_k}$ to update during training. In other words, we never need to train 3D filter coefficients, but just the weights of linear combination. \par 
Here, we assume $\mathbf{F}$ represent a random matrix whose columns   $\mathbf{f_i},1\leq i \leq Q$, are orthonormal random vectors of dimension $LD^2 \times 1$ so that $\mathbf{F^TF} = \mathbf{I}$ is the identity matrix. The question is how should such random vectors be chosen to represent the underlying knowledge space of the filters?  Of course, the ideal (but unknown) filter $\mathbf{h_k}$  (which are of size $LD^2 \times 1$) can be exactly represented as a linear combination of $LD^2$ such orthogonal random vectors. However, since $\mathbf{F}$ only has $Q$ columns, the error between the ideal filter and its linear approximation $\mathbf{\hat{h}_k} = \mathbf{Fb_k}$ is

\begin{equation}
    \mathbf{e} = \mathbf{Fb_k-h_k}
\end{equation}

The minimum squared error solution is  $\mathbf{b_k}=\mathbf{F^T h_k}$, which yields

\begin{equation}
    \mathbf{||e||^2} = \mathbf{h^T_k}[\mathbf{FF^T-I}]\mathbf{h_k}
\end{equation}

Therefore, the \textit{relative error} is bounded by

\begin{equation}
    (1 - \lambda_{max}) \leq \frac{||\mathbf{e}||_2}{||\mathbf{h_k}||_2} \leq (1 - \lambda_{min}) 
\end{equation}

 where $\lambda_{min}$, and $\lambda_{max}$ are minimum and maximum eigenvalues of $\mathbf{FF^T}$. In other words, an upper bound on the relative approximation error can be minimized by making $\lambda_{min}$ as large as possible, while the lower bound can be reduced by ensuring that $\lambda_{max}$ is also large as possible.  The sample realizations of the random vectors used for actual experiments can be judiciously chosen to achieve these objectives to ensure that they serve as reasonable choice for basis filters.

\section{Background and Related Work}
Filter pruning is probably the earliest explored research directions for compression and efficient implementation of CNNs. L. Cun et al. \cite{optimalbrain} and Hassibi et al. \cite{Surgeon} showed that second derivative of the loss can be used to reduced the number of connections in a network. This strategy not only yields an efficient network but also improves generalization. However these methods are only applicable for training the network from scratch. More recently however there has been growing interest in pruning redundancies from a pre-trained network. Han et al. \cite{Han2015LearningBW} proposed a compression method which aims to learn not only weights but also the connections between neurons from training data. While Srinivas et. al. \cite{Srinivas2015DatafreePP} proposed a data-free method to prune neurons instead of the whole filters. Chen et al. \cite{Chen2015CompressingNN} proposed a hash based parameter sharing strategy which intern reduces storage requirements. \par

Filter quantization has been also used for network compression. These methods aim to reduces the number of bits required to represent the filters which can in turn lead to efficient CNN implementation. Quantization using k-means clustering has been explored by Gong et al. \cite{Gong2014CompressingDC} and Wu et al. \cite{Wu2016QuantizedCN}. Similarly Vanhoucke et al. \cite{Vanhoucke2011ImprovingTS} also showed that 8-bit quantization of the parameters can result in significant speed-up with minimal loss of accuracy. In contrast \cite{Han2016DeepCC} combined quantization with pruning. A special case of quantized networks are binary networks, which use only one bit to represent the filter values. Some of the works which explore this direction are BinaryConnect \cite{Courbariaux2015BinaryConnectTD}, BinaryNet \cite{Courbariaux2016BinaryNet} and XNORNetworks \cite{Rastegari2016XNORNetIC}. Their main idea is to directly learn binary weights or activation during the model training. \par

Knowledge Distillation methods train a smaller network to mimic the output(s) of a larger pre-trained network. \cite{Bucila2006ModelC} is one of the earliest works exploring this idea. They trained a smaller model from a complex ensemble of classifiers without significant loss in accuracy. More recently \cite{Hinton2015DistillingTK} further developed this method and proposed a knowledge distillation framework, which eased the training of networks. Another  adaption of \cite{Bucila2006ModelC} is  \cite{Ba2014DoDN} which aims to compress deep and wide networks into shallower ones. \cite{Belagiannis2018AdversarialNC} also used this idea to transfer knowledge from larger networks to much shallower ones using adversarial loss. 

oOur work relates to a class of techniques that rely on basis functions to represent the convolution filters, but differs in several key respects. For instance in \cite{Jaderberg2014SpeedingUC}, Jaderberg et al have proposed a similar two stage decomposition in terms of basis functions followed 1D convolutions for recombining outputs of basis filters. However, to achieve processing speed, their focus is on approximating full rank filter banks using rank-1 filter basis filters, which were optimized to reconstruct the original filters and the response of the CNN to the training data. It was shown that this method leads to significant speed up of a four stage CNN for character recognition. However, the authors do not address the problem of learning in compressed format, nor how this method might impact the performance of other well known CNN architectures on standard data sets.  Qiu et al \cite{Qiu2018DCFNetDN} have also observed that a conventional convolution can be represented as a two successive convolutions involving a basis set and projection coefficients, but their construct differs from the one proposed in Figure 1. Their focus is on 2D Fourier Bessel functions as a basis set for reducing the number of operations required within a given 3D filter kernel, while noting that random basis functions also tend to perform well.  Although this method learns with fewer parameters than conventional CNNs, our approach exploits the redundancy in the full 3D structure of the convolution layer (across all channels and filters) and therefore necessitates even fewer learnable parameters. \par

\section{Experiments}

As described in section 3, we can use BasisConv to compressedly represent all pre-trained convolution layers in a traditional ConvNet. Additionally we can also train such networks (referred to as BasisNet), from scratch in this format. We now describe our experiments in detail.

\subsection{Datasets and Models}
\label{data-and-models}
We performed our experiments on three publicly available image classification datasets. These are \textbf{CIFAR10}, \textbf{CIFAR100} and \textbf{SVHN}. All three datasets contain 32 $\times$ 32 pixel RGB images. CIFAR10 and SVHN contain 10 object classes while CIFAR100 has 100 classes. \par 
We tested  four different CNN architectures with BasisConv. These are \textbf{Alexnet} \cite{Krizhevsky2012ImageNetCW}, \textbf{VGG16} \cite{Simonyan2015VeryDC}, \textbf{Resnet110} \cite{He2016DeepRL} and \textbf{Densenet190} \cite{Huang2017DenselyCC}. We used the pytorch implementation and pre-trained weights of these networks provided by \cite{bearpaw}, since this implementation is suitable for 32 $\times$ 32 input images unlike the originallly proposed networks which are designed for 224 $\times$ 224 input size.

% \vspace{-8pt}
% \paragraph*{SVHN} This a challenging digit classification dataset which contains 10 classes (digits 0 - 9) with 73,257 training images and 26,032 test images. Size of each image is 32 by 32 pixels.
% \vspace{-8pt}
% \paragraph*{CIFAR10} It contains 10 object classes and has 60000 RGB images of 32 by 32 pixel resolution, split into two sets of 50000 training and 10000 test images. 
% \vspace{-8pt}
% \paragraph*{CIFAR100} This is the more difficult version of the CIFAR10 dataset with 100 object classes. Number of images in test and train set are same as CIFAR10 but with 100 classes, it has much less number images per class in training and test set.

\begin{figure}
  \centering
  
  \subfloat[]{\includegraphics[clip, trim=2cm 7cm 3cm 8cm, scale=0.40]{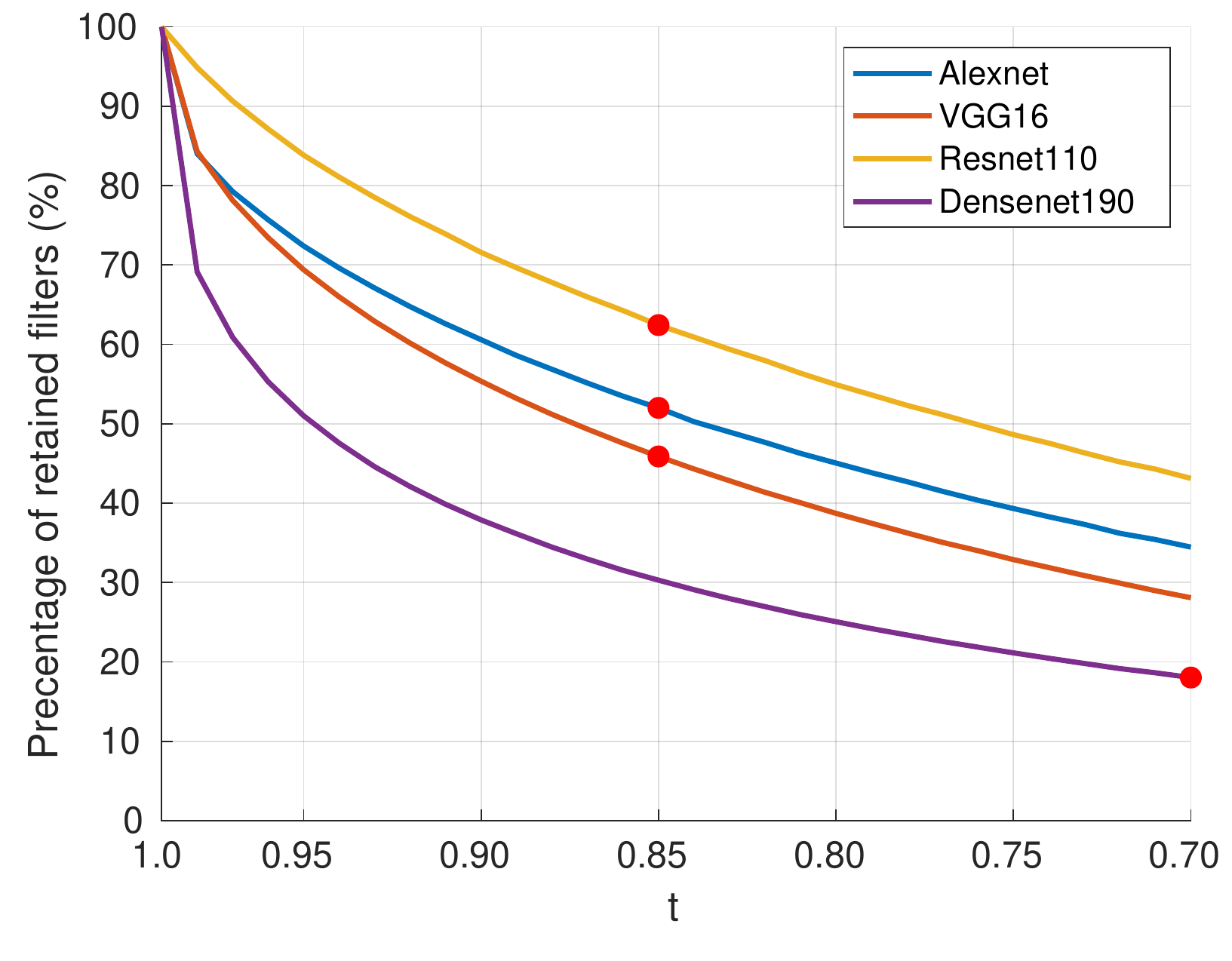}\label{fig:filters-t}}
  \subfloat[]{\includegraphics[clip, trim=2.5cm 7cm 3cm 8cm, scale=0.40]{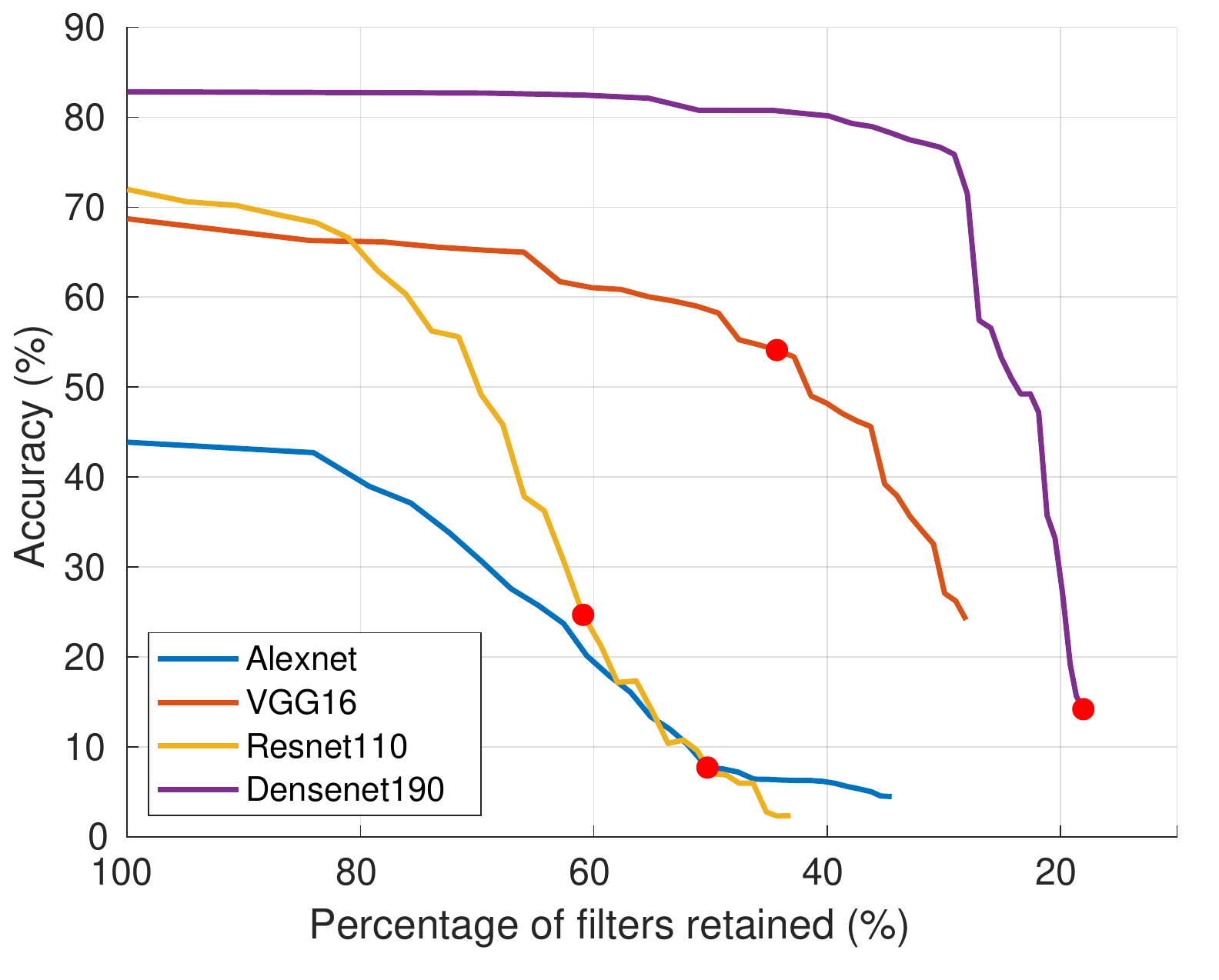}\label{fig:accuracy-filters}}
  \caption{A comparison of compressiblity for 4 network architectures, pre-trained on CIFAR100 dataset. (a) Shows the number retained basis filters as a percentage as we reduce \textbf{t} from 1.0 to 0.7,  while (b) plots the test accuracy against the percentage of retained filters. Red dots indicates initial operating points for the compressed network, which is later fine tuned to improve accuracy to obtain the  results shown in Table \ref{tab:compression-filters} }
  \label{fig:compression-filters}
\end{figure}

% \vspace{-8pt}
% \paragraph*{Network Architectures}
% We tested  four different CNN architectures with BasisConv. These are Alexnet, VGG16, Resnet110 and Densenet190. We used the pytorch implementation of these networks meant for CIFAR dataset, provided by \cite{bearpaw}. We choose this implementation instead of native pytorch implementation of these networks because of the inconsistency of input images size. Pytorch implementation of these networks assume an input image of size 224 x 224 pixels so instead of unnecessarily resizing 32 x 32 pixels images we choose the networks which readily accept 32 x 32 pixel images. We also used the pretrained weights, when possible, or training parameters provided by \cite{bearpaw}.

\begin{table}[htb]
\centering
\caption{Comparison of number of learnable parameters (in Millions) between ConvNet and BasisNet.}
\vspace{3pt}
\label{tab:compression-parameters}
\begin{tabular}{l|llll}
\toprule
\bf Model & \bf t & \bf ConvNet & \bf BasisNet \bf & \scriptsize \textbf{\begin{tabular}[c]{@{}c@{}} Learnable Parameter \\ Reduction Factor \end{tabular}} \\ 
\midrule
Alexnet & 0.85 & 2.4954 & 0.1867 & 13.4\\
VGG16 & 0.85 & 14.766 & 0.8312 & 17.8 \\ 
Resnet110 & 0.85 & 1.7366 & 0.1399 & 12.4 \\ 
Densenet190 & 0.70 & 25.8216 & 1.4363 & 18.0 \\ 
\bottomrule
\end{tabular}
\end{table}

\subsection{Network Compression}
To compress the pretrained network we replaced each convolution layer in the network with the BasisConv layer. The resulting network is refered as BasisNet. BasisConv layer implements two operation, i) convolution with the basis filters and ii) linear combination of output using projection coefficients, which is implemented as convolution with 1x1 filters. Parameters for the BasisConv layer are computed from the weights of original convolution layer using eigen decomposition as explained in section \ref{Compression-of-pretrained-Networks}. Compression emerges from the the fact that only a small number (Q) of basis filters are needed to reconstruct the output of convolution layer and rest of them can be safely discarded. To determine Q, we first sort the eigenvectors such that their corresponding eigenvalues are in descending order.  The first Q eigenvectors are then selected such that the ratio of the sum of their eigenvalues and the sum of all eigenvalues exceeds a threshold \textbf{t}. Naturally maximum value of \textbf{t} is 1.0 at which point each BasisConv layer retains all basis filters and hence all of the information  contained in the original convolution layer. As we reduce \textbf{t} we are able to discard more number of filters corresponding to the smaller eigenvalues with some drop in test accuracy.  Figure \ref{fig:compression-filters} compares the compressed potential of all four networks, pre-trained on CIFAR100. Figure \ref{fig:filters-t} shows the percentage of retained filters in compressed network as we reduce \textbf{t} from 1.0 to 0.7, while figure \ref{fig:accuracy-filters} plots the accuracy against the percentage of retained filters. In this figure we  see that, all four networks can discard 20\% of their filters with little change in test accuracy with Densenet190 being the most compressible which can discard upto 60\% filters with only 3\% drop in accuracy. It should be noted that these plots show the performance of the networks prior to fine tuning of the learnable parameters (also referred to as $\textbf{w}_k$  in Figure 1), and the red dots indicate the compression points selected for  performance optimization by subsequent fine tuning.

% Please add the following required packages to your document preamble:
% \usepackage{multirow}
\begin{table}[htb]
% \small
\centering
\caption{Comparison of maximum compression achieved for each network  architecture pre-trained on the CIFAR100 dataset. As we can see Densenet190 is most compressible with the reduction in number of filters and multiplications by a factor of more than 5 while keeping the accuracy with in 3\% of the original network.}
\vspace{3pt}
\label{tab:compression-filters}
\resizebox{\textwidth}{!}{%
\begin{tabular}{l|lll|lllll}
\toprule
\multicolumn{1}{c}{\multirow{6}{*}{\textbf{Model}}}  & \multicolumn{3}{c|}{\textbf{Original ConvNet}} & \multicolumn{4}{c}{\textbf{Compressed BasisNet}} \\
\cmidrule(r){2-4}
\cmidrule(r){5-9}
\multicolumn{1}{c}{} & \multicolumn{1}{c}{\textbf{\begin{tabular}[c]{@{}c@{}}\# Filters\end{tabular}}} & \multicolumn{1}{c}{\textbf{GFlops  \footnotemark}} & \multicolumn{1}{c|}{\textbf{Accuracy}} & \multicolumn{1}{c}{\textbf{t}} & \multicolumn{1}{c}{\textbf{\begin{tabular}[c]{@{}c@{}} \# Filters\end{tabular}}} & \multicolumn{1}{c}{\textbf{GFlops}} & \multicolumn{1}{c}{\scriptsize \textbf{\begin{tabular}[c]{@{}c@{}} Accuracy \\ Before \\ Finetuning\end{tabular}}} & \multicolumn{1}{c}{\scriptsize \textbf{\begin{tabular}[c]{@{}c@{}} Accuracy \\ After \\ Finetuning\end{tabular}}} \\
\midrule
Alexnet & 1152 & 0.24 & 43.9 \% & 0.85 & 358 & 0.085 & 7.1 \% & 42.5 \% \\
VGG16 & 4187 & 0.313 & 68.7 \% & 0.85 & 1855 & 0.18 & 54.1 \% & 67.3 \% \\
Resnet110 & 4096 & 0.253 & 72.0 \% & 0.85 & 1719 & 0.11 & 24.7 \% & 69.9 \% \\
Densenet190 & 20117 & 18.678 & 82.8 \% & 0.70 & 3525 & 3.5 & 14.2 \% & 80.7 \% \\
\bottomrule
\end{tabular}}
\end{table}
\footnotetext{GFlop refers to number of multiplications in Billions, counting only the multiplications in convolutional layers.}

% % Please add the following required packages to your document preamble:
% % \usepackage{multirow}
% \begin{table}
% \small
% \centering
% \caption{Comparison of maximum compression achieved for each network architecture.}
% \vspace{3pt}
% \label{tab:compression-filters}
% \resizebox{\textwidth}{!}{%
% \begin{tabular}{l||l|l|l|l|l|l|l|l}
% \toprule
% \multicolumn{1}{c}{\multirow{4}{*}{\textbf{Model}}} & \multicolumn{3}{c}{\textbf{Original ConvNet}} & \multicolumn{4}{c}{\textbf{Fine-tuned BasisNet}} \\
% \cmidrule(r){2-4}
% \cmidrule(r){5-9}
% \multicolumn{1}{c}{} & \multicolumn{1}{c}{\textbf{\begin{tabular}[c]{@{}c@{}}\# Filters\end{tabular}}} & \multicolumn{1}{c}{\textbf{GFlops}} & \multicolumn{1}{c}{\textbf{Accuracy}} & \multicolumn{1}{c}{\textbf{t}} & \multicolumn{1}{c}{\textbf{\begin{tabular}[c]{@{}c@{}} \# Filters\end{tabular}}} & \multicolumn{1}{c}{\textbf{GFlops}} & \multicolumn{1}{c}{\textbf{\begin{tabular}[c]{@{}c@{}} asd \\ Filters\end{tabular}}} & asd \\
% \midrule
% Alexnet & 1152 & 0.24 & 43.9 \% & 0.85 & 358 & 0.085 & 42.5 \% \\
% VGG16 & 4187 & 0.313 & 68.7 \% & 0.85 & 1855 & 0.18 & 67.3 \% \\
% Resnet110 & 4096 & 0.253 & 72.0 \% & 0.85 & 1719 & 0.11 & 69.9 \% \\
% Densenet190 & 20117 & 18.678 & 82.8 \% & 0.70 & 3525 & 3.5 & 80.7 \% \\
% \bottomrule
% \end{tabular}}
% \end{table}

\subsection{Fine tuning of learnable parameters} As we further reduce \textbf{t} we are able to get more compression but test accuracy also drops significantly. To mitigate this we train each network in two steps for a total of 25 epochs. In step one we train the projection coefficients (i.e. the 1D filters  $\textbf{w}_k$) only for 15 epochs with SGD. Since our network has significantly less learnable parameters (see Table \ref{tab:compression-parameters}) 15 epochs are enough to re-train these coefficients. We used step learning rate starting with 0.1 and dividing by 10 every 5 epochs. In step two we update all non-convolutional parameters (including the fully connected layers) in the network (but hold the basis filters constant) for another 10 epochs with 5e-4 learning rate. This process enables us to recover test accuracy even when large numbers of basis filters are discarded. Table \ref{tab:compression-filters} compares the the maximum compression we were able to achieve for all four networks, pre-trained on CIFAR100, while keeping the accuracy within 3\% of the original network. We can see here that Densenet190 is the most compressible with reduction in number of filters by a factor of 5.7 and reduction in multiplications by a factor of more than 5.3.

\begin{wrapfigure}{r}{0.5\textwidth}
  \begin{center}
  \vspace{-30pt}  
    \includegraphics[clip, trim=3cm 7cm 3cm 8cm, scale=0.4]{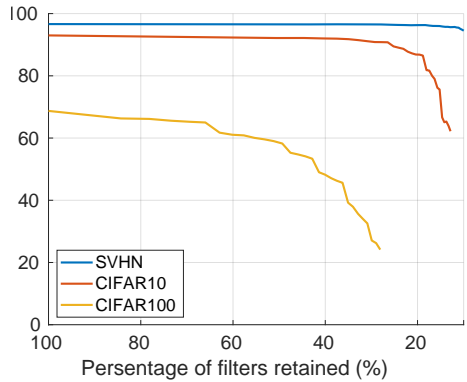}
  \end{center}
  \caption{Test accuracy vs compression, for the VGG16 trained on three datasets}
  \vspace{-10pt}
  \label{tab:accuracy-data}
\end{wrapfigure}
% \begin{wrapfigure}
%   \centering
%   \includegraphics[clip, trim=0cm 0cm 6cm 0cm, scale=0.49]{Figures/accuracy-data.pdf}
%   \caption{}
%   \label{fig:basisconv}
% \end{wrapfigure}

\subsection{Network Compression vs Dataset} Intuitively it is clear that complexity of information learned by a network during training must depend on the complexity of the dataset it was trained on. This means that the same network architecture trained on different datasets will have different compressibility. To verify this, we trained VGG16 on three image classification detests mentioned in section \ref{data-and-models}. Figure \ref{tab:accuracy-data} shows the graph of test accuracy for these datasets plotted against the percentage of filters retained in the compressed network. These trends are consistent with our intuition that the SVHN data set is the simplest and the resulting trained network is highly compressible. On the other hand, the CIFAR100 is the most complex of the three datasets, which is reflected in the faster drop in performance with increasing compression. Not surprisingly, as the complexity of the problem increases, more knowledge is stored in each convolution layer, and larger number of eigenfilters are required to capture the most relevant information.  

\subsection{Training from scratch}
To illustrate the process for learning with random basis sets, we describe an example provided in Matlab 2018b that trains a simple CNN for image classification on CIFAR10 data set.  The original network configuration (shown in Table \ref{tab:net-comparisson} on left) has three convolution layers (two of size 5x5x32 and one of size 5x5x64). This network is trained for 40 epochs, and achieves a classification test accuracy of 74\%. The number of learnable parameters in the three convolution layers is 79,328. On the right, each conventional layer is replaced by the BasisConv structure which reduces the number of learnable parameters to 6400 (i.e. a reduction by a factor of 12). In this configuration, the 3D filters in the layers marked “conv\_1”, “conv\_3”, and “conv\_5” are initialized as orthonormal random functions but then held frozen during the learning process, while the 1D filters marked “conv\_2”, “conv\_4”, and “conv\_6” are allowed to update. After 40 epochs, the configuration on the right achieves a test accuracy of 71\%.  Setting aside the fully connected layers (which are common to both configuration), this experiment illustrates how BasisConv reduces the number of learnable parameter by an order of magnitude, without significant loss in performance. Additionally we also trained Alexnet and VGG16 from scratch with random basis in pytorch. In these experiments we normalized the intermediate tensor with BatchNormalization before convolving with 1D filters, $\mathbf{w_k}$. Recall that the conventional versions of these networks achieve 43.9\% and 68.7\% accuracy on the CIFAR100 data set, respectively. We were able to get 42.5 \% and 66.3 \% using BasisConv for Alexnet and VGG16 respectively, which is within  2\% of the accuracy of original network while reducing the the number of learnable parameters by a factor of 7.2 and 7.9 respectively.

% \begin{table}[]
% \centering
% \caption{}
% \label{tab:my-table}
% \begin{tabular}{llll}
% Model & t & ConvNet & BasisNet \\
% Alexnet & 0.85 & 2.4954 & 0.1867 \\
% VGG16 & 0.85 & 14.766 & 0.8312 \\
% Resnet110 & 0.85 & 1.7366 & 0.1399 \\
% Densenet190 & 0.70 & 25.8216 & 1.4363
% \end{tabular}
% \end{table}

% Please add the following required packages to your document preamble:
% \usepackage{booktabs}
\begin{table}[]
\centering
\caption{A comparison of a conventional convolutional network (left) with the corresponding basis convolution network (right). In BasisConv, the 3D convolution kernels are never trained (indicated by *), which reduces the number of learnable parameters.}
\vspace{3pt}
\label{tab:net-comparisson}
\resizebox{\textwidth}{!}{%
\begin{tabular}{@{}l|llll||ll|lll@{}}
\toprule
Layer & Activations & Weights & Bias & \ & \ &  Layer & Activations & Weights & Bias \\ \cmidrule(r){1-4}
\cmidrule(r){7-10}
Input & 32x32x3 & - & - & \ & \ & Input & 32x32x3 & - & - \\
conv\_1 & 32x32x32 & 5x5x3x32 & 1x1x32 & \ & \ &  conv\_1 & 32x32x32 & 5x5x3x32* & 1x1x32 \\
relu\_1 & 32x32x32 & - & - & \ & \ &  conv\_2 & 32x32x32 & 1x1x3x32 & 1x1x32 \\
maxpool\_1 & 15x15x32 & - & - & \ & \ &  relu\_1 & 32x32x32 & - & - \\
conv\_2 & 15x15x32 & 5x5x32x32 & 1x1x32 & \ & \ &  maxpool\_1 & 15x15x32 & - & - \\
relu\_2 & 15x15x32 & - & - & \ & \ &  conv\_3 & 15x15x32 & 5x5x32x32* & 1x1x32 \\
maxpool\_2 & 7x7x64 & - & - & \ & \ &  conv\_4 & 15x15x32 & 1x1x32x32 & 1x1x32 \\
conv\_3 & 7x7x64 & 5x5x32x64 & 1x1x64 & \ & \ &  relu\_2 & 15x15x32 & - & - \\
relu\_3 & 7x7x64 & - & - & \ & \ &  maxpool\_2 & 7x7x32 & - & - \\
maxpool\_3 & 3x3x64 & - & - & \ & \ &  conv\_5 & 7x7x64 & 5x5x32x64* & 1x1x64 \\
fc\_1 & 1x1x64 & 64x576 & 64x1 & \ & \ &  conv\_6 & 7x7x64 & 1x1x64x64 & 1x1x64 \\
relu\_4 & 1x1x64 & - & - & \ & \ &  relu\_3 & 7x7x64 & - & - \\
fc\_2 & 1x1x10 & 10x64 & 10x1 & \ & \ &  maxpool\_3 & 3x3x64 & - & - \\
softmax & 1x1x10 & - & - & \ & \ &  fc\_1 & 1x1x64 & 64x576 & 64x1 \\
\ & \ & \ & \ & \ & \ &  relu\_4 & 1x1x64 & - & - \\
\ & \ & \ & \ & \ & \ &  fc\_2 & 1x1x10 & 10x64 & 10x1 \\
\ & \ & \ & \ & \ & \ &  softmax & 1x1x10 & - & - \\ \bottomrule
\end{tabular}}
\end{table}
\section{Conclusion}
In summary, we have presented a general method for network compression and efficient implementation which can be easily incorporated into existing CNN architectures. For pre-trained network, each convolution layer is replaced by two successive convolutions: first with  eigen basis filters (that capture the underlying knowledge space of the layer), followed by 1D kernels (that can be finetuned) to generate the activations. We used four network architectures and three datasets to show that our method consistently reduces i) the number of learnable parameters by an order of magnitude, and ii) multiplications and filter storage by as much as a factor of 5, with less than 3\% degradation in performance. Finally, using random basis functions and significantly fewer learnable parameters, BasisNet achieve comparable performance to a conventional CNNs when learning from scratch.
\mbox{}
\nocite{*}
\bibliographystyle{unsrt}
\bibliography{neurips_2019}

\end{document}